\documentclass[runningheads]{llncs}
\usepackage[T1]{fontenc}
\usepackage{graphicx}
\usepackage{adjustbox}
\usepackage{booktabs}
\usepackage{amsmath}
\usepackage{amssymb}
\begin{document}
\title{Explainable Multimodal AI Approach to Reduce Unnecessary Breast Biopsies}
%
%
\author{
Sadam Hussain\inst{1}\orcidID{0000-0002-3453-0785}\thanks{Corresponding author: a01753094@exatec.tec.mx}
\and
Mansoor Ali\inst{1}\orcidID{0000-0002-4669-7267}
\and
Daly Betzabeth Avendaño-Avalos\inst{2}\orcidID{0000-0002-8433-7211}
\and
Jose Gerardo Tamez Peña\inst{2}\orcidID{0000-0003-1361-5162}
}
%
%
\institute{
School of Engineering and Sciences, Tecnologico de Monterrey, Monterrey, Nuevo Leon, Mexico \\ \email{a01753094@exatec.tec.mx}
\and
School of Medicine and Health Sciences, Tecnologico de Monterrey, Monterrey, Nuevo Leon, Mexico
}

 \maketitle              
\begin{abstract}
Recently, there has been rise in multimodal vision-language models for computer-aided diagnosis (CAD), yet their application in breast diagnosis, particularly for predicting biopsy outcomes remains limited. This study presents a multimodal vision-language approach, incorporating Grad-CAM as an Explainable AI (XAI) technique, to predict benign tumors in patients recommended for biopsy, especially those classified as BI-RADS 4 and 5. A novel multimodal framework was developed using Vision Transformer (ViT) for imaging feature extraction and \texttt{DeBERTa-V3-Large} for textual embeddings. The model was trained and evaluated on a comprehensive in-house dataset comprising 1,632 patients, each with four mammography views and a corresponding radiology report. The features of both modalities were extracted independently and fused by late fusion before classification. Grad-CAM was used to generate visual explanations highlighting key regions influencing model predictions.
The proposed model demonstrated high sensitivity 90\%, 95\%, 99\%, and 100\% with corresponding specificities of 40\%, 36\%, 18\%, and 11\%, respectively. The best AUC achieved was 0.72 by multimodal approach on the test set, outperforming the image-only (0.66) and text-only (0.68) baselines. The proposed multimodal architecture, enhanced with Grad-CAM, demonstrates a strong potential to identify benign tumors among high-risk patients, offering interpretable insights and thereby helping reduce unnecessary biopsies, patient anxiety, and healthcare costs.

\end{abstract}
\section{Introduction}
Breast cancer is the most common disease among women in the United States and the world \cite{siegel2023cancer}. Different imaging modalities have been used to diagnose breast cancer in the early stage. Mammography is one of the common imaging techniques used for earlier cancer diagnosis and has shown potential benefits in clinical settings \cite{siegel2023cancer}. One of the main problems with mammography is that most benign findings are recommended for biopsy. BI-RADS was created to describe the various stages of cancer appropriately to provide better patient care \cite{lacquement1999positive}. BI-RADS category 4—is divided into four subsections; 4, 4a, 4b, and 4c, represents a suspicious abnormality, which is therefore recommended for biopsy. Among the recommended biopsies in the BI-RADS 4 category in general, 75\% of the BI-RADS category 4 findings are proven to be benign \cite{lacquement1999positive}. In the United States alone, more than 970,000 breast biopsies are unnecessary \cite{brentnall2018long}. This causes a high false positive rate, which leads to additional expenditure, clinical procedures, and further stress.
Different CAD systems have been used to improve the diagnosis and prognosis of breast cancer in various stages using numerous imaging modalities and clinical data \cite{shen2017deep}. The latest breakthroughs in machine learning (ML) and deep learning (DL) models have enabled the implementation of CAD models \cite{lecun2015deep}. The ever-burgeoning capacity of computational resources and access to a large amount of data creates extraordinarily favorable conditions for DL models in various image classification, segmentation, and object detection tasks and enables extracting insight from textual data in parallel \cite{yala2021toward}. DL models use Convolutional Neural Networks (CNNs) to automatically extract imaging features, in contrast to conventional hand-crafted techniques \cite{lecun2015deep}. CNN-based advanced architectures have shown potential in different breast cancer diagnosis and prognosis applications \cite{wang2016deep}, such as breast cancer classification \cite{lotter2021robust}, risk prediction \cite{yala2021toward}, survival \cite{arya2021multi}, and recurrence prediction \cite{massafra2021clinical}. In recent studies, a surge in the use of vision-language models for different applications in classification, detection, and segmentation has been observed in biomedical and general-purpose applications \cite{yan2022clinical}. However, it is rare in breast cancer diagnosis and prognosis applications due to the lack of multimodal medical data.
\newline
BI-RADS 4 and associated categories of patients are recommended for biopsy. Few studies have used DL approaches to predict biopsy outcomes \cite{meng2023reducing}; however, to the best of our knowledge, no study has employed a multimodal approach combined with an XAI technique for this purpose. Therefore, a novel multimodal framework that employs a ViT to extract mammographic features and \texttt{DeBERTa-V3-Large} to obtain embeddings from radiological reports is proposed. Furthermore, an XAI technique—Grad-CAM, is used to visualize the image regions that contribute to the model's prediction for interpretability purposes. ViT has demonstrated effective performance in tumor classification, and Grad-CAM—an XAI technique, is applied to generate attention-based heatmaps. Grad-CAM highlights the critical imaging regions that contribute the most to the model’s predictions, thereby offering interpretability and fostering trust in automated decision-making systems \cite{selvaraju2017grad}.

\noindent In the proposed framework, features from each modality are processed independently and subsequently fused using a late-fusion strategy through feature concatenation before the final classification layer. This multimodal integration aims to address the limitations of unimodal approaches by incorporating both mammographic imaging data and radiological reports. Four mammographic views—right cranial-caudal (R-CC), the right mediolateral-oblique (R-MLO), the left cranial-caudal (L-CC), and the left mediolateral-oblique (L-MLO), are used as inputs to the model. The objective of this study is to use a vision–language modeling approach to assess biopsy outcomes for BI-RADS 4 and its subcategories, as well as BI-RADS 5, with the goal of reducing unnecessary biopsies for patients who do not have breast cancer.

\section{Related work}

The classification of benign tumors from breast biopsy recommended that patients can be seen in ultrasound imaging studies using ML. Investigators in a few studies \cite{zhao2020reducing}, have explored the possibilities of identifying benign biopsy from ultrasound images using ML methods. However, it is uncommon to classify biopsy-recommended patients in mammography images using ML techniques. Only a handful of studies have used these techniques to investigate this problem further. An unnecessary breast biopsy reduction problem has been explored using DL on digital mammograms quite recently. Meng and colleagues \cite{zhao2020reducing} explored the possibility of classifying patients based on BI-RADS 4 using a deep transfer learning technique. They particularly used a pre-trained Inception-V3 model. The best AUC reported for this model is 0.86. The authors in a study \cite{wu2021reducing}, developed a patch-based DL pipeline, where they divided mammography images into different patches to obtain local features. They aggregated feature maps using a shallow convolutional aggregation network, which includes two convolutional layers, global average pooling, and a classification layer. They used DenseNet-161 for extracting local features and reported an AUC of 0.799. In another study \cite{ezeana2023deep}, authors designed a decision support tool for risk categorization, and reduction of unnecessary biopsies in mammographic data. An AUC of 0.93 is reported in this study. In a recent study conducted by Liu and colleagues \cite{liu2024deep}, used a simple DL VGG16 architecture on digital mammograms to classify benign tumors among malignant tumors in the category 4 BI-RADS. They reported the best AUC of 0.72 for the main task of classifying benign tumors among various types of tumors, along with a few other complementary tasks. 

\noindent Limited research has been conducted on the identification of the biopsy outcome using ML from BI-RADS 4 patients. Few of the studies that have used DL to predict biopsy outcomes from BI-RADS 4 patients are presented as follows. The authors \cite{meng2023reducing} used deep transfer learning on digital mammograms for patients with BI-RADS 4 recommended for biopsy. An Inception-V3 model was particularly used for this objective. In another study \cite{wu2021reducing}, the authors extracted patches from mammography images to capture important local features. They aggregated these patches using a shallow convolutional aggregation network and subsequently fed the aggregated feature maps into a DenseNet-161 architecture for final outcome classification. In another work \cite{ezeana2023deep}, researchers designed a decision support tool used to stratify risk and reduce biopsy in BI-RADS 4 mammograms. The study was limited in the sense that the model was built and validated internally. In a recent preliminary study \cite{liu2024deep} investigators designed a DL model based on digital mammograms to avoid unnecessary breast biopsies. This study performed four binary classification tasks—the main task being the classification of tumors as benign versus all other types of lesions, achieving a best AUC score of 0.72. However, the study had limitations, such as a limited internal dataset consisting only of mammograms and using a simple VGG16 architecture. 
\noindent In summary, existing studies on biopsy outcome prediction from BI-RADS 4 mammograms remain limited in several respects. Most approaches rely solely on mammographic images and do not leverage the complementary information contained in radiological reports. Furthermore, many studies are developed and evaluated using relatively small, single-institution datasets, which may limit generalizability. In addition, model interpretability has received limited attention, reducing transparency and potentially hindering clinical adoption. These limitations highlight the need for multimodal and explainable approaches for biopsy outcome prediction.

\section{Methodology}
This section provides a detailed description of the dataset used, processing methods, and data augmentation approach applied on the data; the proposed vision-language architecture; and implementation details. The architecture involves integrating four-view mammography (LCC, LMLO, RCC, RMLO) and radiological reports using the late fusion technique.

\subsection{Dataset Overview}
\begin{figure}[t]
\centering
\includegraphics[width=\linewidth, keepaspectratio]{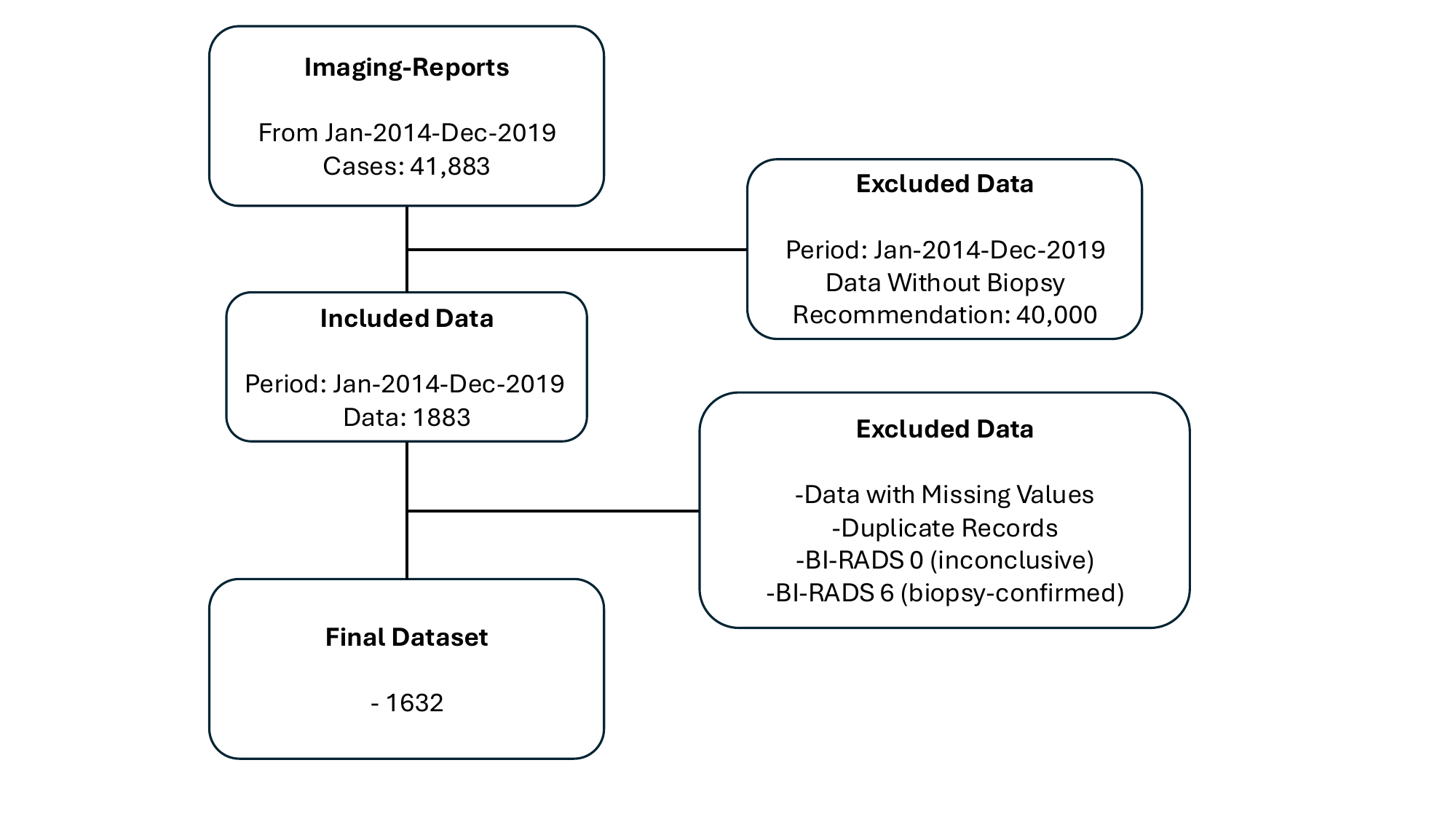}
\caption{Dataset inclusion-exclusion criteria for false positive biopsy reduction multimodal model.}
\label{fig4}
\end{figure}

\begin{figure}[t]
\centering
\includegraphics[width=\linewidth, keepaspectratio]{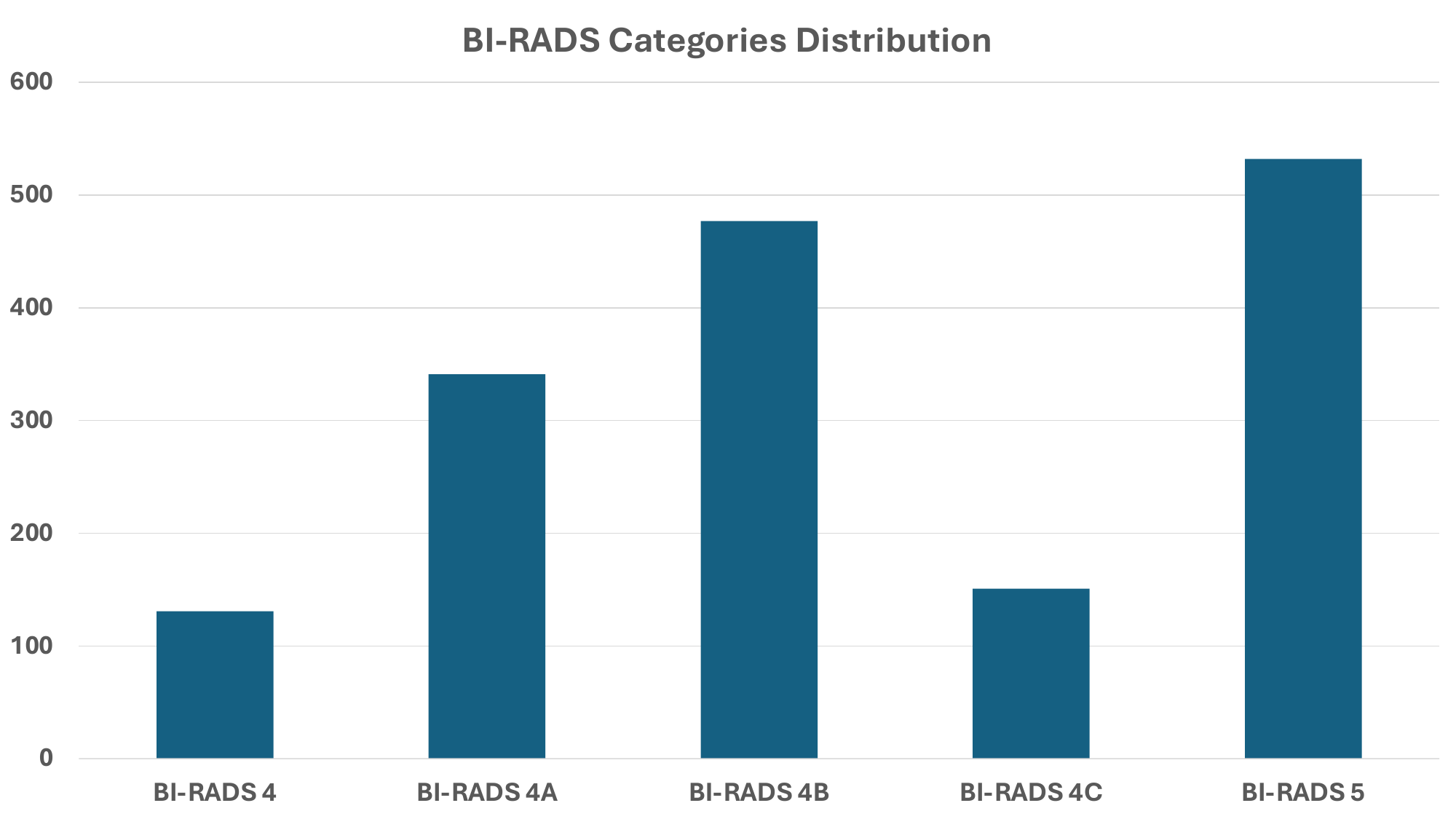}
\caption{The data distribution of BI-RADS categories.}
\label{fig3}
\end{figure}

The TecSalud hospital's at Tecnologico de Monterrey provided the dataset for this study. Together with digital mammograms, it also comprises digital breast tomosynthesis (DBT), digital mammography (DM), and ultrasound (US) based radiological reports. All reports and mammograms were anonymized to maintain privacy. The reports were gathered between 2014 and 2019, originally written in Spanish. Google Translate was used to translate them into the English, and bilingual radiologists then confirmed the translation. Two experienced radiologists wrote and assessed each report. The dataset started with 42000 examples. Only biopsy-recommended patient data were extracted. The four views (LCC, LMLO, RCC, and RMLO), as well as the corresponding radiological reports containing detailed descriptions of the diagnosis—including BI-RADS classifications, breast density scores, patient ages, family cancer histories, lesion laterality, biopsy recommendations, among others were combined with each patient's data. The laterality of the lesion shows whether the patient has ever had breast cancer. If a patient had cancer in either breast, it is indicated by the laterality of the lesion. Three values are used to describe laterality: 0, 1, and 2. Cancer in the left breast is represented by the number 0, cancer in the right breast by the number 1, and cancer in both breasts by the number 2. The purpose of laterality was to determine whether the patient had cancer at all. Furthermore, 1,632 cases comprising a total of 6,528 mammograms, along with their associated radiological reports, were included after preprocessing. The preprocessing steps involved removing reports with missing values, duplicate records, and cases assigned BI-RADS categories 0 (incomplete assessment), 1 (negative), 2 (benign finding), 3 (probably benign), and 6 (biopsy-proven malignancy). Thus, the dataset contains the remaining categories of BI-RADS 4 and associated categories 4A, 4B, 4C, and BI-RADS 5. The distribution of BI-RADS used in this study can be seen in the Fig. \ref{fig3}. 
The patients, who were all female and used for the final model building and assessment, had an average age of 51.52, a standard deviation (SD) of 11.44, and an age range of 22-93. 
The report is a sample and can be used for demonstration purpose. The detailed characteristics of the dataset are defined in Table \ref{table1}. 
The inclusion-exclusion criteria of the data used in this experiment can be seen in the Fig. \ref{fig4}.


\begin{table*}[t]
\centering
\caption{Characteristics of the dataset.}
\label{table1}
\begin{tabular}{p{5.5cm} p{7cm}}
\hline
\textbf{Characteristics of Dataset} & \textbf{Details} \\
\hline
No. of Examples & 1632 \\
Sex & Women (1632), 100\% \\
Mexican Cohort & North East Mexico \\
Age (mean), SD, Range & 51.52, 11.44, 22-93 \\
Biopsy Recommendation & 1632 \\
Family History of Cancer & No: 1448, Yes: 184 \\
Breast Density Distribution (ACR) & 
Fatty: 62\\
& Scattered fibroglandular tissue: 415\\
& Heterogeneously dense: 793\\
& Extremely dense: 362 \\
Distribution of BI-RADS & 
4: 131\\
& 4A: 341\\
& 4B: 477\\
& 4C: 151\\
& 5: 532 \\
\hline
\end{tabular}
\end{table*}

\begin{figure*}[t]
\centering
\includegraphics[width=0.8\linewidth, keepaspectratio]{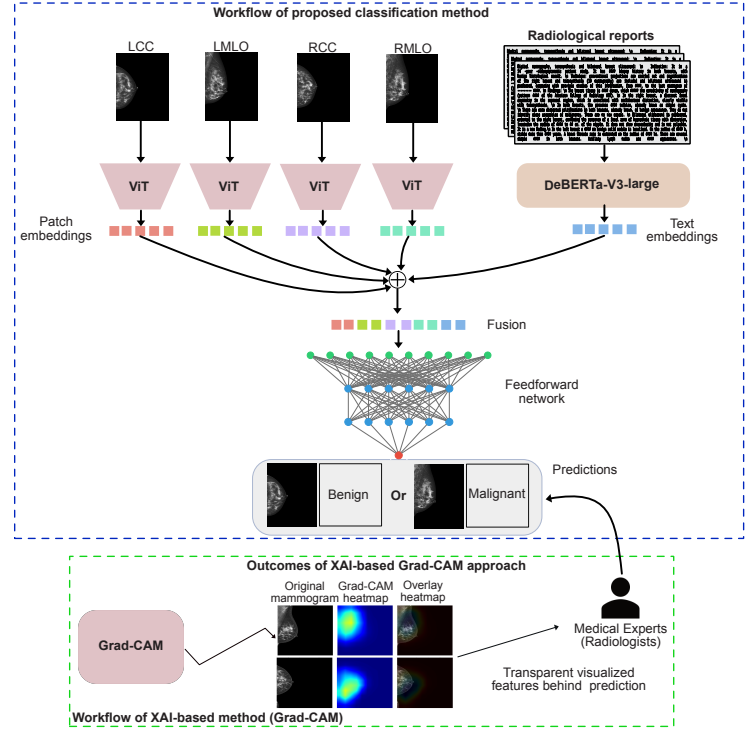}
\caption{In the first box, the proposed multimodal approach used for breast biopsy outcome predictions is represented. ViT \cite{dosovitskiy2021an} was used to extract embeddings from four views (R-CC, R-MLO, L-CC, L-MLO) of digital mammograms and the \texttt{DeBERTa-V3-Large} \cite{he2023debertav} model to extract textual embeddings from radiological reports. Afterwards, late feature fusion technique to combine features from both modalities for the final biopsy outcome classification was used. The second box represents the GRAD-CAM applied to mammograms to explain the model decision and subsequently verified by the radiologist.}
\label{workflow}
\end{figure*}

\subsection{Data Preprocessing}
Each patient's record had a concise clinical indication, personal risk history, imaging descriptions, findings, and diagnostic impressions derived from the radiological reports. The BI-RADS classifications, breast density scores, patient ages, and family cancer histories were collected from the reports. These extracted data were used as input to the multi-modal architecture, along with multi-view mammograms.

\noindent The mammography dataset was subjected to various augmentation approaches, meticulously selected to bring diversity and resilience into the training material. Both horizontal and vertical mirroring were used to display the mammograms in various orientations to the model. In addition, a random scaled cropping technique was applied to ensure that the model remains invariant to changes in scale and translation. The cropping was specifically targeted to achieve a size of 224x224 pixels. In addition, dynamic modifications were incorporated into the images, including shifting, scaling, and rotating. The rotations ranged from 0 to 90 degrees, while the scaling factor varied from 0.8 to 1.2 times the original size. These diverse augmentation techniques successfully enlarged the training dataset. The augmentation strategy in this study introduced the model to several views, which prevents overfitting and improves the model's ability to handle different input types, thus enhancing its overall performance.

\subsection{Model Architecture}
This study introduces a multimodal DL architecture aimed at classifying or predicting biopsy outcomes for BI-RADS 4 and 5 lesions. The model employs distinct DL components to extract features from mammography images and textual data. In the imaging component, a pre-trained ViT architecture \cite{dosovitskiy2021an}, specifically using the \texttt{ViT-base-patch16-224-in21k} model was used to improve the feature extraction capability of the network. This model is pre-trained on \texttt{ImageNet-21k} (14 million images, 21,843 classes) at a resolution of 224x224. The ViT processes input mammography images and leverages its powerful attention mechanisms to capture intricate details and relationships within the images, complementing the classification task. 
In this architecture, the initial layers of the ViT were frozen to retain the pre-trained features and unfroze the last 11 layers to allow fine-tuning on the biopsy outcome prediction task. This approach enables the model to leverage both pre-trained knowledge and task-specific learning for improved classification performance.
The ViT processes the mammography images to extract detailed feature representations. These image features are then fused with textual features extracted from clinical reports.
\newline
A variant of DeBERTa \cite{he2021deberta}, \texttt{DeBERTa-V3-Large} \cite{he2023debertav} is used to extract embeddings from clinical reports. The \texttt{DeBERTa-V3-Large} model comprises 24 layers with a hidden size of 1024. It contains approximately 304 million backbone parameters and an embedding layer with 131 million parameters, derived from a vocabulary of 128,000 tokens. DeBERTa\_V3 is pre-trained on large-scale general-domain corpora using a replaced token detection objective, enabling it to capture rich contextual and semantic representations from clinical text. It is subsequently fine-tuned to better adapt to domain-specific textual inputs. The extracted image features from the ViT, with a dimension of 768, are concatenated with the textual features from \texttt{DeBERTa-V3-Large}, resulting in a unified multimodal feature vector. This vector is then passed through a linear classification layer to predict biopsy outcomes. By integrating ViT and \texttt{DeBERTa-V3-Large}, the model improves its ability to represent and distinguish relevant multimodal patterns, enabling robust analysis for breast cancer diagnosis.


\noindent This method enables the separate extraction of mammography and textual characteristics, which are subsequently combined through a late feature fusion mechanism. More precisely, the ViT extracts mammography features, combined with the textual features processed by the \texttt{DeBERTa-V3-Large}. The combined characteristics are then entered into an Artificial Neural Network (ANN) structure for the ultimate categorization assignment. This integrated methodology capitalizes on the advantages of utilizing both visual and textual data, facilitating a more resilient and all-encompassing examination to differentiate between non-cancerous and cancerous tumors. The workflow is depicted in Fig. \ref{workflow}.

\noindent In this study, a late fusion strategy was employed to integrate features from multiple modalities before classification. The integration process involved concatenating features extracted from different sources—mammograms and radiological reports. The details of a late feature fusion approach and classification process are mentioned below.

\subsubsection{Feature Extraction:}
Let $\mathcal{F}_{\text{img}} \in \mathbb{R}^{d_{\text{img}}}$ represent the feature vector extracted from the image modality using a ViT, where $d_{\text{img}} = 768$. Let $\mathcal{F}_{\text{text}} \in \mathbb{R}^{d_{\text{text}}}$ denote the feature vector extracted from the radiological reports using \texttt{DeBERTa-V3-Large} model, where $d_{\text{text}} = 1024$.

\subsubsection{Feature Concatenation:}
The combined feature vector $\mathcal{F}_{\text{concatenated}}$ is obtained by concatenating $\mathcal{F}_{\text{img}}$ and $\mathcal{F}_{\text{text}}$:

\begin{equation}
    \mathcal{F}_{\text{concatenated}} = concat[\mathcal{F}_{\text{img}}, \mathcal{F}_{\text{text}}].
\end{equation}

Therefore, the dimensionality of $\mathcal{F}_{\text{concatenated}}$ is given by:

\begin{equation}
d_{\text{combined}} = d_{\text{img}} + d_{\text{text}} = 768 + 1024 = 1792.    
\end{equation}

\subsubsection{Classification:}
The final prediction $\hat{y}$ is computed by applying a linear transformation to the concatenated feature vector followed by a softmax activation function. Formally, let $\mathbf{W} \in \mathbb{R}^{C \times d_{\text{combined}}}$ be the weight matrix and $\mathbf{b} \in \mathbb{R}^C$ be the bias vector, where $C$ represents the number of classes. The prediction $\hat{y}$ is given by:

\begin{equation}
\hat{y} = \text{softmax}(\mathbf{W} \cdot \mathcal{F}_{\text{concatenated}} + \mathbf{b}),    
\end{equation}

where $\text{softmax}$ is the activation function that converts the linear output into class probabilities.

\subsection{Transparency and Integration of Grad-CAM in Breast Cancer Diagnosis:}

\subsubsection{Transparency in Decision-Making}
One of the primary challenges of implementing AI in healthcare is the "black-box" nature of complex models, which often lack interpretability. Grad-CAM helps address this by introducing transparency into decision-making. In the context of breast cancer diagnosis and prognosis, this transparency is crucial, as it enables researchers and clinicians to understand the basis of the system's predictions. By generating heatmaps, Grad-CAM provides intuitive visual aids that highlight critical regions in mammograms, thus bridging the gap between complex AI models and their practical applications in healthcare.

\subsubsection{Integration of Grad-CAM for Enhanced Interpretability:}
To improve the model’s transparency and assist medical professionals in making accurate decisions, Grad-CAM is integrated with outcomes from linear support vector machines. The method generates heatmaps that emphasize the regions in imaging data—such as mammograms—that the model focuses on while predicting tumor presence. The Grad-CAM heatmap is computed as follows:

\begin{equation}
    \text{Grad-CAM}(i, j) = \text{ReLU} \left( \sum_k \alpha_k A^k_{i,j} \right),
\end{equation}

where $A^k_{i,j}$ represents the activation maps of the final convolutional layer at spatial location $(i, j)$, and $\alpha_k$ is the importance weight for feature map $k$, calculated as:

\begin{equation}
    \alpha_k = \frac{1}{Z} \sum_i \sum_j \frac{\partial y^c}{\partial A^k_{i,j}},
\end{equation}

where $y^c$ is the class score for class $c$, and $Z$ is the normalization factor over the spatial dimensions.

To further enhance interpretability, an integrated explanation is constructed by combining the original mammogram with the Grad-CAM heatmap. This is expressed as:

\begin{equation}
    IE = \beta \cdot \text{Original Image} + (1 - \beta) \cdot FH,
\end{equation}

where $IE$ is the integrated explanation, $FH$ is the Grad-CAM-generated heatmap, and $\beta$ is a weighting factor controlling the contribution of the original image and the heatmap. 

\noindent By incorporating Grad-CAM into the diagnostic workflow, clinicians gain valuable visual explanations of AI predictions, fostering trust in AI-driven models. The proposed framework supports the development of automated systems for breast cancer diagnosis, ensuring interpretable and reliable outcomes to aid in smart healthcare.

\subsection{Implementation Details}
An NVIDIA RTX A4000 GPU with 16 GB memory, PCI Express Gen 4, and third-generation tensor CUDA cores for both training and testing all models were used. The proposed network architecture utilized a ViT backbone to extract features from mammograms, while \texttt{DeBERTa-V3-Large} was used to extract characteristics from textual data. The models underwent training using pre-existing weights. The Adam optimizer was employed with a learning rate of 0.0005, and a batch size of 4 was used. The training process involved training the models for a total of 30 epochs. The activation function utilized was the Rectified Linear Unit (ReLU), with a dropout rate of 0.2. The loss function employed was Binary Cross Entropy (BCE) with logits. The formula for this loss is provided in equation \ref{eq1}. The dataset was partitioned into training, validation, and testing sets, with proportions of 60\%, 20\%, and 20\% accordingly. The evaluation of model performance was conducted using the testing set.

\begin{equation}
\mathcal{L}_{bce}= -\frac{1}{N} \sum_{i=1}^{N} y_i \cdot \log(p(y_i)) + (1 - y_i) \cdot \log (1-p(yi)).
\label{eq1}
\end{equation}

\section{Results}

A retrospective dataset of 1632 patients, which totals 6528 mammograms (four views per patient) and 1632 complete radiological reports were used. The dataset included 131 (8.02\%) BI-RADS 4 cases, 341 (20.89\%) BI-RADS 4A cases, 477 (29.22\%) BI-RADS 4B cases, and 151 (9.25\%) BI-RADS 4C cases, as well as 532 (32.59\%) BI-RADS 5 cases. The mean age of patients was 51.52 years with a standard deviation of 11.44 years, and the age range was 22 to 93 years.

\noindent In proposed architecture for mammography feature extraction, the initial layers of the ViT were frozen to retain the pre-trained features and unfroze the last 11 layers to allow fine-tuning on the specific task. This approach enables the model to leverage both pre-trained knowledge and task-specific learning for improved classification performance. Additionally, the \texttt{DeBERTa-V3-Large} model for processing the radiological reports was used—extracting textual features to complement the image features.

\noindent To ensure model interpretability, Grad-CAM was applied to generate heatmaps, visualizing the specific regions of mammography images that most influenced the predictions. This added layer of explainability facilitates understanding of the model's decision-making process, which is particularly important in clinical applications where trust and transparency are critical. These heatmaps can also aid clinicians in identifying imaging patterns relevant to biopsy outcome predictions, further enhancing the model’s usability.

\begin{figure}[t]
\centering
\includegraphics[width=\linewidth, keepaspectratio]{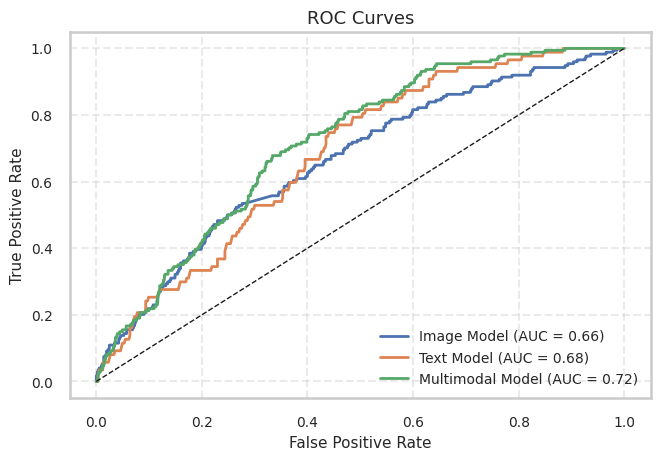}
\caption{An ROC curve of the image-only, text-only, and multimodal models that represent the ratio between true positive rate and false positive rate.}
\label{fig2}
\end{figure}

\noindent The proposed multimodal classifier, which combines image features from mammograms with textual data from radiological reports, achieved an accuracy score of 87\%. The highest precision obtained was 94\%, indicating a high rate of correctly identified positive cases. The best AUC score for predicting biopsy outcome was 0.72, achieved on the testing dataset, which constitutes 20\% of the total dataset. The ROC curve representing the ratio between true positive rate and false positive rate can be seen in Fig. \ref{fig2}. This AUC value demonstrates the model's ability to distinguish between positive and negative biopsy outcomes effectively.

\noindent The image-only model demonstrated modest specificity improvements across increasing sensitivity thresholds, with specificity scores of 1.0\%, 3.0\%, 11.0\%, and 25.0\% at fixed sensitivities of 100\%, 99\%, 95\%, and 90\%, respectively. This indicates a limited reduction in unnecessary biopsies when using imaging data alone. In contrast, the text-only model achieved higher specificity values of 12.0\%, 15.0\%, 24.0\%, and 37.0\% at the same sensitivity levels, suggesting a greater capability to reduce false positives from radiology reports alone.

\noindent The multimodal model consistently outperformed both unimodal approaches, reaching specificity scores of 11.0\%, 18.0\%, 36.0\%, and 40.0\% at 100\%, 99\%, 95\%, and 90\% sensitivity, respectively. These findings highlight that up to 40\% of unnecessary biopsies could be avoided while maintaining 90\% cancer detection, demonstrating the potential of multimodal learning to enhance clinical decision-making. A summary of these results, along with AUC scores, is provided in Table~\ref{table3}.


\noindent Overall, proposed model's performance demonstrates a significant advancement in the biopsy outcome predictions of breast cancer patients, providing a robust tool for clinicians to enhance decision-making processes. The integration of multimodal data, utilizing both the ViT for image features and the \texttt{DeBERTa-V3-Large} model for textual features, the fine-tuning of these models, and the use of Grad-CAM for interpretability were crucial in achieving these results. These contributions pave the way for future research and potential clinical applications. 


\begin{table*}[t]
\centering
\caption{Specificity (\%) at fixed sensitivity levels and AUC scores for each model. Higher specificity at high sensitivity indicates better reduction of false positives while maintaining diagnostic coverage.}
\label{table3}
\resizebox{\textwidth}{!}{%
\begin{tabular}{lcccccc}
\toprule
\textbf{Model} & \textbf{100\% Sens.} & \textbf{99\% Sens.} & \textbf{95\% Sens.} & \textbf{90\% Sens.} & \textbf{AUC} \\
\midrule
Image Model(ViT)       & 1.3  & 3.4  & 11.2 & 25.1 & 0.663 \\
Text Model(DeBERTa)        & 11.6 & 15.3 & 24.3 & 37.0 & 0.681 \\
Multimodal Model  & 11.2 & 17.7 & 35.8 & 40.4 & 0.721 \\
\bottomrule
\end{tabular}%
}
\end{table*}

\section{Discussion}
In this work, a multimodal DL and LLM-based model was developed that integrates mammography images and radiological reports. The objective of this study was to create an automated DL pipeline that can help categorize BI-RADS 4 patients as benign, with the goal of reducing unnecessary breast biopsies—thereby alleviating patient anxiety, physical discomfort from biopsy procedures, and associated healthcare costs. To this end, a dataset comprising patients with BI-RADS 4 and 5 assessments, covering subcategories 4, 4A, 4B, and 4C were collected. A tumor classification workflow with direct clinical implications, was built and evaluated across a range of diagnostic sensitivity levels.

\noindent In addition to the multimodal model, image-only and text-only models were independently trained and evaluated. While the image-only model demonstrated limited specificity improvements, the text-only model achieved better performance in reducing false positives when used in isolation. This supports the hypothesis that clinical context captured in radiology reports complements image features, and motivates the integration of both modalities. As a novel multimodal approach, this study builds upon previous unimodal research \cite{liu2024deep}, and demonstrates that SOTA DL models can effectively leverage digital mammograms and radiological narratives to support decision-making in BI-RADS 4 assessments. The proposed multimodal model achieved strong diagnostic performance and showed potential to reduce unnecessary biopsies by 40\%, 36\%, 18\%, and 11\% while maintaining sensitivities at 90\%, 95\%, 99\%, and 100\%, respectively.

\noindent To ensure interpretability and enhance clinical relevance, XAI techniques like Grad-CAM were applied to generate heatmaps that visually identify the specific regions of mammography images that significantly influenced the model's predictions. These heatmaps not only improve transparency but also provide radiologists with insights into the decision-making process of the model. The application of Grad-CAM in this pipeline bridges the gap between black-box DL models and practical clinical implementation, enabling clinicians to trust and validate the model’s predictions. Additionally, these visual explanations can aid in highlighting imaging patterns relevant to distinguishing benign tumors from malignant ones, further reinforcing the usability of proposed approach in clinical settings.

\noindent Although proposed multimodal approach produces superior performance in terms of disease detection, resulting in reducing unnecessary biopsies, there are a few limitations as well. First, the data is limited and obtained from a single institute, which could limit this approach for generalization on other datasets. Second, the dataset is collected locally and therefore, represents a single race; thus, the characteristics of the patients can be limited to a single race. Data from different institutes should be obtained to further validate proposed multimodal approach for better generalization. Third, radiologists in practice use different imaging modalities, such as breast ultrasound, contrast-enhanced mammography (CEM), computed tomography (CT), and magnetic resonance imaging (MRI) images, among others, for evaluation~\cite{steyerova2022current}. On the contrary, this approach only includes mammograms. Fourth, while Grad-CAM improves model transparency and facilitates clinical interpretability, other advanced XAI techniques could be explored to further elucidate the model's reasoning. Lastly, pre-trained vision-language models were used; however, different attention-based mechanisms can be applied to extract rich contextual features from images and text, which can improve the overall performance.

\noindent In the future, it is planned to extend the dataset and add other modalities and multidimensional data to improve current model's performance further. The integration of more advanced XAI techniques alongside Grad-CAM will also be investigated to provide a more comprehensive explanation of model predictions and enhance clinical adoption.

\section{Conclusion}

In conclusion, a novel SOTA multimodal vision-language model was developed to classify biopsy outcomes in patients with BI-RADS 4 and its subcategories. Leveraging the strengths of multimodal learning in CAD, digital mammograms processed using ViT and radiological reports encoded with \texttt{DeBERTa-V3-Large} were combined. Features extracted were fused for classification. Proposed model outperformed unimodal baselines, achieving an AUC of 0.72, accuracy of 87\%, and precision of 94\%. At sensitivities of 90\%, 95\%, 99\%, and 100\%, it yielded specificities of 40\%, 36\%, 18\%, and 11\%, respectively, indicating up to 40\% reduction in false positive biopsies. Grad-CAM visualizations enhanced interpretability, aiding clinical decision-making. This initial multimodal approach demonstrates strong potential to reduce unnecessary biopsies, offering clinically meaningful benefits in diagnostic accuracy, patient reassurance, and cost reduction.

\section{Acknowledgment}
This research was supported by the Secretaría de Ciencia, Humanidades, Tecnología e Innovación (Secihti) and Tecnologico de Monterrey.

%
%
%
\bibliographystyle{splncs04}
\bibliography{ref}

\end{document}